\documentclass[11pt,a4paper,dvipsnames]{article}
\usepackage{emnlp2021}
\usepackage{hyperref}
\usepackage{times}
\usepackage{latexsym}

\usepackage{float}
\usepackage[utf8]{inputenc}
\usepackage[T1]{fontenc}
\usepackage{tikz}
\usepackage{amssymb}
\usepackage{amsmath}
\usepackage{multirow}
\usepackage{array}
\usepackage{graphicx}
\usepackage{microtype}

\usepackage{algorithm2e} 

\usepackage{linguex} 

\title{Uncovering More Shallow Heuristics: Probing the Natural Language Inference Capacities of Transformer-Based Pre-Trained Language Models Using Syllogistic Patterns}

\author{Reto Gubelmann \\
  University of St.Gallen \\
  Rosenbergstrasse 30 \\
  9000 St.Gallen\\
  \texttt{reto.gubelmann@unisg.ch}\\
  \And
  Siegfried Handschuh \\
  University of St.Gallen\\
  Rosenbergstrasse 30 \\
  9000 St.Gallen \\ 
\texttt{siegfried.handschuh@unisg.ch}\\ 
}
\date{}

\begin{document}
\maketitle

\begin{abstract}
In this article, we explore the shallow heuristics used by transformer-based pre-trained language models (PLMs) that are fine-tuned for natural language inference (NLI). To do so, we construct or own dataset based on syllogistic, and we evaluate a number of models' performance on our dataset. We find evidence that the models rely heavily on certain shallow heuristics, picking up on symmetries and asymmetries between premise and hypothesis. We suggest that the lack of generalization observable in our study, which is becoming a topic of lively debate in the field, means that the PLMs are currently not learning NLI, but rather spurious heuristics.

\end{abstract}


\section{Introduction}

Current natural language inference (NLI) is typically conceived as a three-way classification problem, taking samples such as \ref{ex:informallyvalid}, consisting of a premise (P) and a hypothesis (H), and requiring the models to categorize them as either \emph{contradiction} (P and H cannot both be true), \emph{entailment} (If P is true, H must be true as well), or \emph{neutral} (neither of the two, that is, given the truth of P, H may or may not be true; this is the case with example \ref{ex:informallyvalid}).

\ex. (P) The streets are wet. (H) It has rained.
\label{ex:informallyvalid}

Transformer-Based pre-trained language models (PLMs) have become the \emph{de facto} standard in a variety of natural language understanding (NLU) tasks, including NLI. Based on the encoding part of the original transformer architecture \citep{Vaswani:2017}, researchers have proposed a number of highly successful NLU architectures, such as BERT \citep{Devlin:2019}, RoBERTa \citep{Liu:2019}, XLNet \citep{Yang:2019}, DeBERTa \citep{he2020deberta}, and smaller versions such as DistilBERT and DistilRoBERTa \citep{Sanh:2019}, MiniLM \citep{Wang:2020} and Albert \citep{lan2019albert}. Additionally, a number of sequence-to-sequence architectures have been proposed that are more similar to the original transformer than to BERT in that they directly try to transform one sequence to another, much like the basic set-up of neural machine translation. These include T5 \citep{Raffel:2019} and BART \citep{Lewis2020BARTDS}. These PLMs are now regularly outperforming the human benchmark, as evinced by the \href{https://gluebenchmark.com/leaderboard}{GLUE} and \href{https://super.gluebenchmark.com/leaderboard}{SuperGLUE} Leaderboards, \citep{Wang:2018} and \citep{Wang:2019}. 

While it is impossible to deny the performance of these models at such benchmarks, it is another question whether this performance is driven by simple shallow heuristics or by any real understanding of the tasks that they are performing. Indeed, there is a growing consensus in the literature that purely neural NLI approaches suffer from the use of shallow heuristics and, as consequence, a lack of generalization beyond the fine-tuning dataset, see \citet{zhou2020towards}, \citet{bras2020adversarial}, \citet{utama2020towards}, \citet{asael2021generative}, \citet{he2019unlearn}, \citet{mahabadi2019end}, and \citet{bernardy2019kind}. 

This study contributes to this analysis of shallow heuristics used by state of the art models using the systematic of syllogistic figures and moods, which marks the beginning of formal logic in the west, but which has not been put to use in current NLI to expose shallow heuristics. We develop a synthetic test dataset based on this systematic and use it to evaluate the n NLI capacities of transformer-based PLMs that have been fine-tuned on the common NLI datasets, in particular SNLI \citep{bowman2015large} and MNLI \citep{Williams2018multinli}.

We begin by reviewing the stat of the art (section 2), then we discuss both the MNLI and the SNLI datasets as well as our own synthetic dataset built on syllogistic (section 3). We then evaluate a number of PLMs that have been fine-tuned on the former datasets and discuss the results (sections 4 \& 5).

\section{Previous Research}

\citet{geirhos2020shortcut} have proposed a general diagnosis of the problem of shallow heuristics, and \citet{Ribeiro:2020} have urged a more comprehensive, multi-dimensional approach to testing the abilities of these models instead of simply submitting them to automated benchmarks. \citet{niven2019probing} consider the almost-human performance of BERT at the Argument Reasoning comprehension task, finding that the models exploit ``spurious statistical cues in the dataset'' to reach this performance; they develop a dataset similar to the original one at which the PLMs do not perform better than random.  

With regard to the main datasets, \citet{gururangan2018annotation} show that SNLI and, to a lesser extent, MNLI, contain cues that make it possible to achieve very good accuracy in categorizing hypotheses by only looking at the hypotheses. \citet{hossain2020analysis} show that simply ignoring negation does not substantially decrease model performance in many NLI datasets. \citet{bernardy2019kind} argue that both SNLI and MNLI only cover a part of the entire range of human reasoning. In particular, they suggest that they do not cover quantifiers, nor strict logical inference. 

With regard to PLMs fine-tuned on these datasets, \citet{wallace-etal-2019-universal} show that certain triggers, inserted context-independently, lead to a stark decline in NLI accuracy. \citet{morris2020textattack} provide a systematic framework to create adversarial attacks for NLU. \citet{Chien2020AdversarialAO} focus on syntactic biases for models fine-tuned on SNLI and MNLI, also finding that these biases are strong.  

\citet{mccoy2019right} hypothesize that state-of-the-art NNLP models use three kinds of heuristic at NLI tasks: the lexical overlap heuristics (which focuses on the overall lexical overlap between premise and hypothesis), the subsequence heuristics (focusing on hypotheses that are a subsequence of supposed premises), and constituent heuristic (focusing on syntactically discrete units of the premise as hypothesis). Using their dataset called HANS that is designed so that any use of these three heuristics results in mistaken predictions, they find that state-of-the-art models in NLI do make many such mistakes, suggesting that they are indeed using these heuristics.

\citet{richardson2020probing}, finally, use cleverly chosen semantic fragments (i.e. subsets of a language translatable into formal logic, in particular first-order predicate logic) to test the models' understanding of the logical relationships of contradiction, entailment and neutral. They find that the models tested perform poorly on these tasks, but that this performance can be remedied with fine-tuning the models on sufficient amounts of training data that has been synthetically generated from these fragments.

In non-automated logical analysis of language, the dominant approach proceeds by using predicate logic, as it has been pioneered by \citet{frege1892sinn} and developed by \citet{russell1905denoting}. For a recent contribution in this tradition with a focus on computability, see \citet{moot2019natural}. We will rely on an older way to formalize logical relationships: syllogistic. It dates back to \citet{analprior}, and it has been somewhat sidelined by predicate logic recently. However, it has one staunch proponent in \citet{Oderberg:2005}. For our purposes, what makes syllogistic so attractive is that, as we shall see, it allows to synthesize large amounts of valid and invalid samples. This is because syllogistic aims at systematically finding all valid inference patterns containing three predicates (a subject-, middle- and predicate-term), which are then combined via quantifiers into three statements (a conclusion and two premises).

\section{Datasets}

\subsection{The SNLI \& MNLI Datasets}

Given the importance of fine-tuning for the entire method as it is currently practiced, it is clear that this method is squarely based on the availability -- and quality -- of large NLI datasets.

The main datasets in use currently for neural NLI are the Stanford Natural Language Inference corpus (SNLI, \cite{bowman2015large}) and the Multi Genre Natural Language Inference corpus (MNLI, \cite{Williams2018multinli}). The main difference between the two corpora is that SNLI derives exclusively from image captions, while MNLI is sourced from 10 different text genres (the two corpora are approximately the same size of some 500k samples). The following discussion will therefore focus on MNLI, being basically a more diverse and (by design) more challenging version of SNLI. 

\citet{Williams2018multinli} have presented crowdworkers with some 433k statements serving as premises, and then asked them to write up one sentence that is entailed by this premise, one that contradicts it, and one that is neutral to it. The instructions given to the crowdworkers are given in full in the appendix. To ensure maximal diversity of styles, the premises originate from ten different genres. \citet[1114-5]{Williams2018multinli} emphasizes that only minimal preprocessing has occurred, filtering duplicates within a genre, sentences with less than eight characters, and manually removing non-narrative writing such as formulae. 

In the following, we highlight two important aspects of the dataset.
 
\paragraph{Alogical Premises} A consequence of the diversity of genres and the near-absence of preprocessing in MNLI is that the corpus contains premises such as \ref{ex:mnli-quest}.

\ex. iuh-huh how about any matching programs
\label{ex:mnli-quest}

It is incoherent to say that questions entail any other statements: to entail something, a statement has to have determinate truth conditions; questions are textbook cases of sentences that have no determinate truth conditions. Questions might presuppose other statements for their appropriateness (asking ``Is it going to rain today?'' is inappropriate if the question is uttered in pouring rain), and they might implicate other statements (asking ``Are you hungry?'' typically implicates that one is ready to offer some food). As a consequence, premises such as \ref{ex:mnli-quest} cannot entail or contradict any other statements; they are alogical. Therefore, every premise-hypothesis-pair that contains \ref{ex:mnli-quest} as a premise should be labelled \emph{neutral}, as the relationship is neither entailment nor contradiction. 

While this might be a rather extreme case, the basic problem is inseparable from the goal of the MNLI dataset, namely to accurately ``represent the full range of American English'' \citep[1114]{Williams2018multinli}. By randomly choosing statements from genres such as conversations, one inevitably ends up with a number of statements that do not ``describe a situation or event'' in the typical sense, indeed, that are alogical insofar as they cannot stand in entailment or contradiction relations to other statements. 

However, as the crowdworkers were instructed to write sentences resulting in entailment and contradiction pairs for each premise without exception, this leads to numerous pairs that should be labelled neutral but are in fact labelled contradiction or entailment. 

We therefore hypothesize that a model fine-tuned on such a dataset would struggle to recognize neutral pairs correctly, as it has been fed with neutral pairs that have wrongly been labelled as contradiction or entailment.

\paragraph{Negation Bias} It is well-documented that the dataset has a negation bias. This is pointed out by \citet{Williams2018multinli} themselves: in 48\% of cases involving a negation, the correct label is contradiction. It is very likely that this bias results from crowdworker tactics: There is no more efficient way to create a sentence that contradicts any given premise by simply negating the premise.

A model fine-tuned on a dataset with such a negation bias would be expected to wrongly label pairs as contradictions just because either the hypothesis or the premise, but not both, contain a negation. The same effect is not to be expected if both premise and hypothesis contain negations, as such samples would not be expected to have been encountered too often during training: unlike a negation in premise or hypothesis, negations in both of them cannot be the consequence of crowdworkers' taking a shortcut when creating contradiction pairs.

\subsection{The Syllogistic Dataset}

While it has so far not been used to assess NLI capacities of NLU models, the systematic behind our dataset dates back to Aristotle. In his \emph{Prior Analytics} (composed around 350 BC), \citet[book 1]{analprior} diligently analyzes the possible combinations of subject-, predicate-, and middle-term \emph{via} quantifiers and negations to form a number of formally valid inferences. He deduces 24 formally valid patterns of inferences, so-called syllogisms. For instance, consider the three sentences \ref{ex:barbara1}, \ref{ex:barbara2}, and \ref{ex:barbara3}.

\ex. All residents of California are residents of the USA. \label{ex:barbara1}

\ex. All residents of Los Angeles are residents of California. \label{ex:barbara2}

\ex. All residents of Los Angeles are residents of the USA. \label{ex:barbara3}

These three sentences together form a formally valid inference: If you accept \ref{ex:barbara1} and \ref{ex:barbara2} as true, on pain of self-contradiction, you must also accept \ref{ex:barbara3} as true. In the systematic of syllogistic, it is a mood of the first figure that goes by the name of ``BARBARA'', the capital ``A'' signifying affirmative general assertions (``All X are Y.''). Note that the individual truth of either one of the sentences is entirely irrelevant for the formal validity of the inference.

Now, consider the formal logical relationship between \ref{ex:contradiction1} and \ref{ex:contradiction2} on the one hand and \ref{ex:contradiction3} on the other hand. By changing one single word, three letters in total, we have switched the relationship from entailment to contradiction: it is not possible that  \ref{ex:contradiction1}, \ref{ex:contradiction2}, and \ref{ex:contradiction3} are all true.

\ex. All residents of California are residents of the USA. \label{ex:contradiction1}

\ex. All residents of Los Angeles are residents of California. \label{ex:contradiction2}

\ex. No residents of Los Angeles are residents of the USA. \label{ex:contradiction3}

Finally, consider the formal logical relationship between \ref{ex:neutral1} and \ref{ex:neutral2} on the one hand and \ref{ex:neutral3} on the other hand. By changing one word, four letters, we switched the relationship from entailment to neutral: If \ref{ex:neutral1} and \ref{ex:neutral2} are both true, \ref{ex:neutral3} may or may not be true.

\ex. All residents of California are residents of the USA. \label{ex:neutral1}

\ex. Some residents of Los Angeles are residents of California. \label{ex:neutral2}

\ex. All residents of Los Angeles are residents of the USA. \label{ex:neutral3}

We are using a total of 12 formally valid syllogisms -- called BARBARA, CELARENT, DARII, FERIO, CESARE, CAMESTRES, FESTINO, BAROCO, DISAMIS, DATISI, BOCARDO, FERISON -- and we manually develop 24 patterns that are very similar to these 12 syllogisms, but where the first and the second sentence together contradict or are neutral to the third sentence. This yields a total of 36 patterns, 12 of which are valid syllogisms, 12 are contradictory, and 12 are neutral. To fit the premise-hypothesis structure expected by the models, we combine premise one and two to form a single premise.

We then use a pre-compiled list of occupations, hobbies, and nationalities to fill the subject- middle- and predicate-terms in these patterns. Using 15 of each of them and combining them with the 36 pattern yields 121500 test cases in total, each consisting of a premise and a hypothesis. For a fully specified sample that instantiates BARBARA, a mood of the first figure, see example \ref{ex:fullyspecified}.

\ex. (P) All Gabonese are Budget analysts, and all Element collectors are Gabonese. (H) All Element collectors are Budget analysts.

\section{Experiment}

We run a total of seven models on our test dataset, all of which are fine-tuned on standard NLI datasets, namely SNLI and MNLI (see table \ref{tab:multinli} for details:  PLMs marked with one star ``*'' have only been fine-tuned on MNLI, PLMs marked with two stars have been fine-tuned on both SNLI and MNLI). The models are provided by \href{Huggingface}{https://huggingface.co} \citep{Huggingface}, three of them by textattack \citep{morris2020textattack}, and four by Cross Encoder \citep{Reimers:2019}. 

The models' performances on MNLI, per our own evaluation (not all of the models provide evaluation scores, and we did not find precise documentation on how the scores were obtained), are given in table \ref{tab:multinli}, for details of the evaluation, see the appendix, section \ref{sec:app-evaluation}.

\begin{table}
  \small
  \centering
  \begin{tabular}[h]{|p{3cm}|l|l|l} Modelname & N-Par. & MNLI-Matched (Acc.)\\
    \hline
    textattack-facebook\-bart-large-MNLI* & 406M & 0.8887\\ 
    nli-crossencoder-deberta-base** & 123M & 0.8824\\
    cross-encoder-nli-roberta-base** & 123M & 0.8733\\
    cross-encoder-nli\-MiniLM2-L6-H768** & 66M &  0.86602\\
    textattack-bert-base-uncased-MNLI* & 109M & 0.8458\\
    nli-crossencoder-distilroberta-base** & 82M & 0.8364\\
    textattack-distilbert\-base-uncased-MNLI* & 66M & 0.8133\\
  \end{tabular}
  \caption{Performance of the models in focus on the MNLI-Matched validation set. PLMs marked with one star ``*'' have only been fine-tuned on MNLI, PLMs marked with two stars have been fine-tuned on both SNLI and MNLI.}
  \label{tab:multinli}
\end{table}

The basic idea behind the experiment is to assess whether the PLMs' performance on our dataset reveals any shallow heuristics learned by the models during fine-tuning on MNLI and SNLI. 

\section{Results}

\begin{figure}
  \centering
  \includegraphics[scale=0.32]{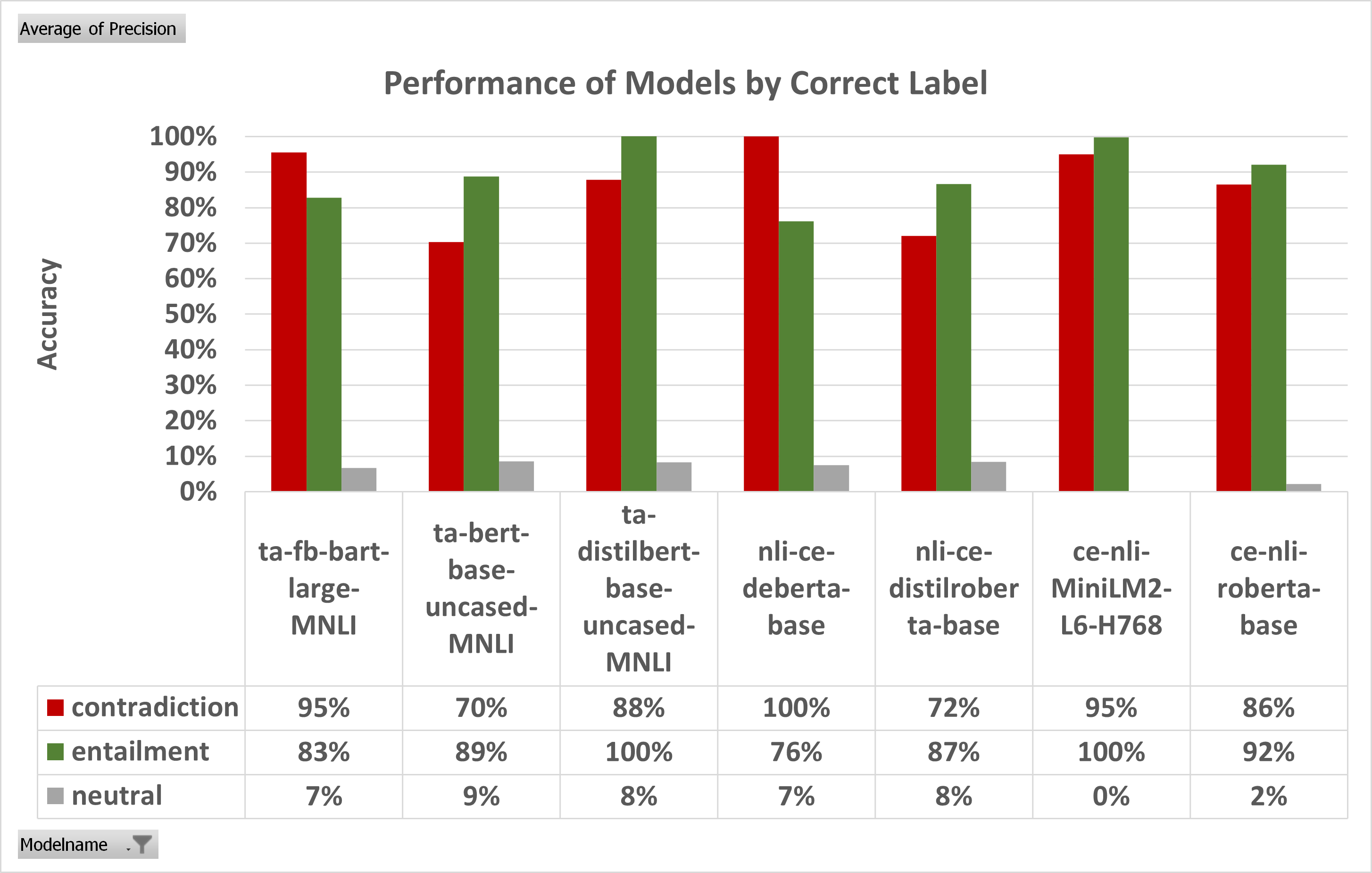}
  \caption{Performance on our syllogistic dataset.}
  \label{fig:results}
\end{figure}

The results of our experiments are shown in figure \ref{fig:results}. For instance, the model whose performance is represented on the very left, textattack's fine-tuned version of BART large, predicts the correct label in only 7\% of cases for neutral labels, while doing so in 95\% for entailment samples and still 83\% for neutral labels.

Figure \ref{fig:results} shows clearly that the models' predictions are quite accurate for labels \emph{entailment} and \emph{contradiction}, but very poor for \emph{neutral}. In table \ref{tab:neutraltop3}, we therefore focus on the top three models' performance for \emph{neutral} samples. The table shows that the two smaller models have a preference for entailment, while the largest model tested has a slight preference for contradiction regarding the neutral models. None of the models achieves accuracy of more than 10\%, which is far below pure guessing (33.3\%).

\begin{table}
  \small
  \begin{tabular}[h]{p{3cm}|l|l|l} Modelname & contrad. & entailm. & neutral\\
    \hline
    textattack-distilbert\-base-uncased-MNLI & 23.68 & 68.01 & 8.31\\
    textattack-facebook\-bart-large-MNLI & 50.64 & 42.67 & 6.69\\
    cross-encoder-nli\-MiniLM2-L6-H768 & 35.41 & 64.49 & 0.1\\
  \end{tabular}
  \caption{For true label neutral, this table gives the percentages of predicted labels for our three best-performing models. For instance, textattack's distilbert erroneously predicts contradiction in 23.68\% of all neutral pairs.}
  \label{tab:neutraltop3}
\end{table}

\section{Discussion}

Overall, textattack's distilbert leads the field with a accuracy of 65\%, which might surprising just because it was among the smallest models evaluated here. However, there is growing evidence that NLI cannot be solved by simply increasing model size. Researchers at DeepMind find that larger models tend to generalize worse, not better, when it comes to tasks involving logical relationships. The large study by \citet[23]{gopher} strongly suggests that, in the words of the authors, ``the benefits of scale are nonuniform'', and that logical and mathematical reasoning does not improve when scaling up to the gigantic size of Gopher, a model having 280B parameters (in contrast, Gopher sets a new SOTA with many other NLU tasks such as RACE-h and RACE-m, where it outperforms GPT-3 by some 25\% in accuracy). 

On the face of it, 65\% looks like an excellent score for a model that has not been trained on the rather challenging dataset that we are using. However, when looking closer at the data, we find that the model performs very poorly with neutral samples; indeed, none of the models is able to recognize such neutral relationships with a accuracy of more than 10\%. Given that pure chance would still yield an accuracy of some 33\%, this is a very poor performance.

\begin{figure}
  \centering
  \includegraphics[scale=0.55]{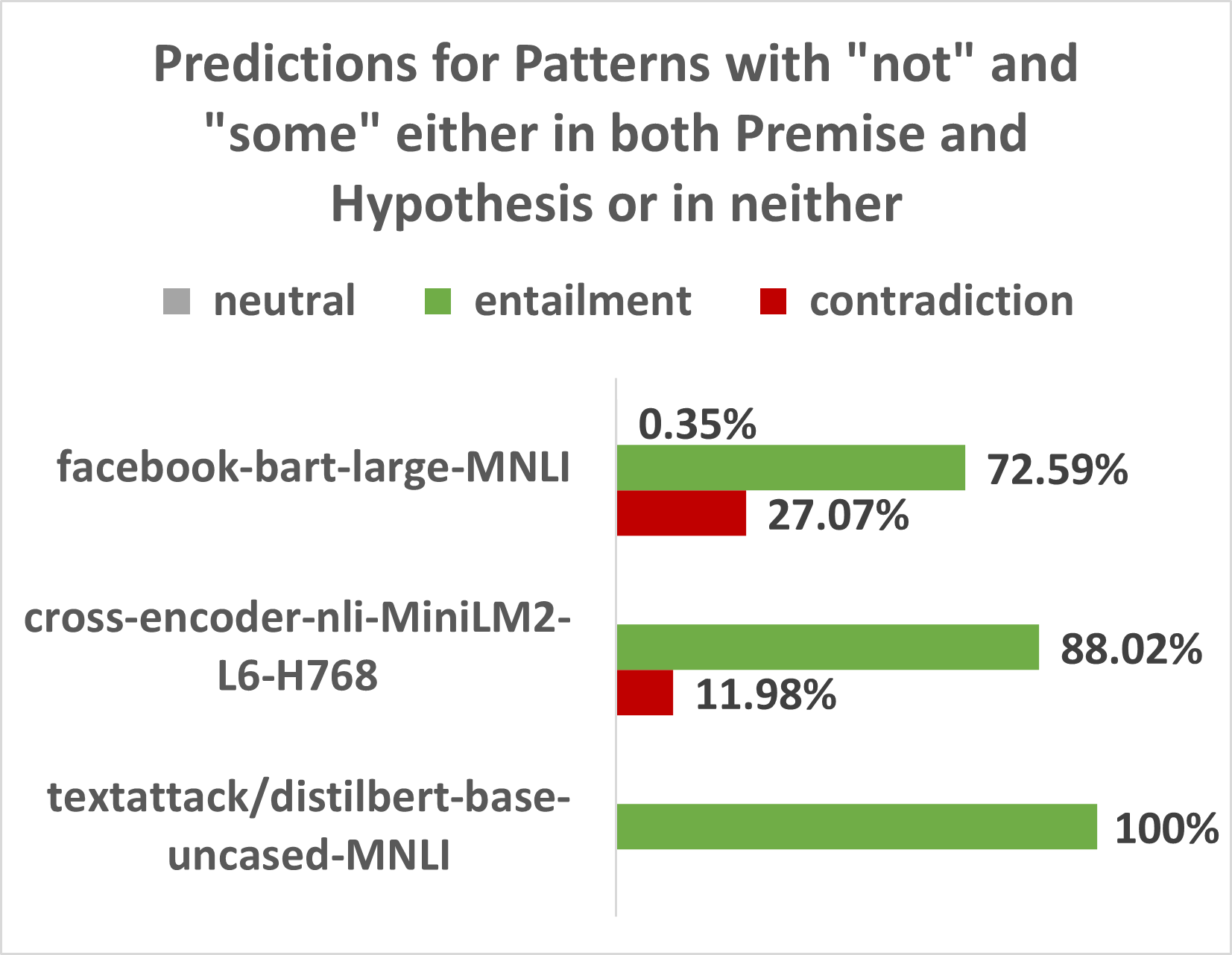}
  \caption{Predicted labels for patterns that are symmetric between premise and hypothesis regarding existential quantifier and negation.}
  \label{fig:symmsomeneg}
\end{figure}

We have therefore further probed the heuristics that the models might be using that could cause the poor performance with neutral labels. Manual inspection showed that they respond strongly to symmetries regarding quantifiers and negations between premises and hypotheses. In particular, if either both or none of the premise and the hypothesis contain a ``some'' (existential quantifier) or a negation (the symmetric conditions), then the models are strongly biased to predict \emph{entailment} (see figure \ref{fig:symmsomeneg}). 

\begin{figure}
  \centering
  \includegraphics[scale=0.55]{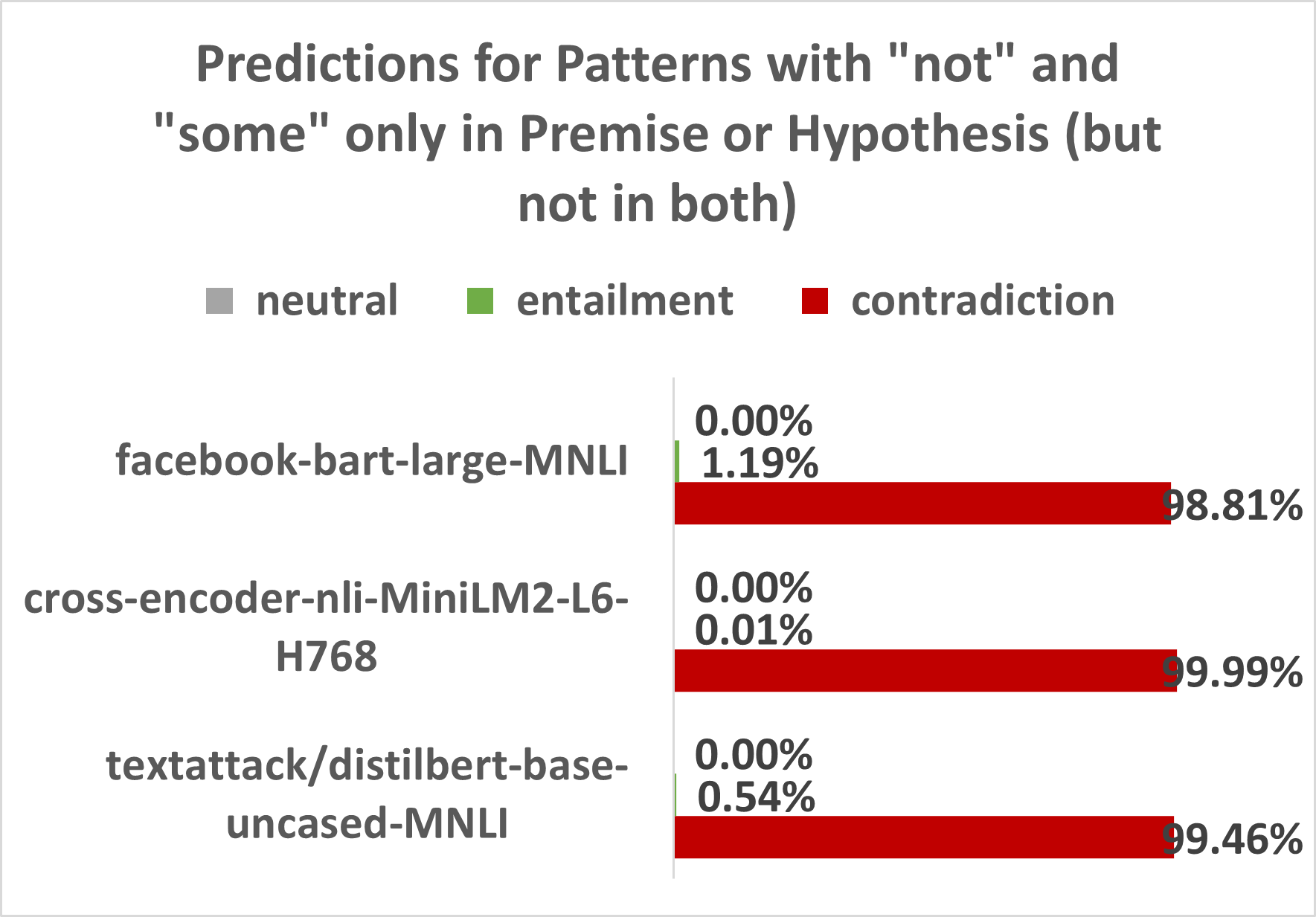}
  \caption{Predicted labels for patterns that are asymmetric between premise and hypothesis regarding existential quantifier and negation.}
  \label{fig:xorsomeneg}
\end{figure}

Conversely, if the pattern contains an asymmetry regarding existential quantifier and negation between premise and hypothesis, then the models are very strongly inclined to predict contradiction (see figure \ref{fig:xorsomeneg}). 

We have been surprised to see that the models are sensitive not only to symmetry regarding the negation particle, but also  regarding the quantifier. 

In the case of the contradiction and entailment pairs, these heuristics serve the models very well in our dataset, resulting in impressive performance. However, when applied to the neutral samples, the heuristics break down, performance falls far below simple guessing.

The observations made regarding the MNLI dataset (see above, section 3.1), together with the overlap biases observed by \citet{mccoy2019right} can to some extent help to explain the models' behavior with neutral samples. Our observations regarding the features of the MNLI dataset, the alogical premises and the negation bias, suggest that a model fine-tuned on this dataset would struggle to identify neutral samples as such: If alogical premises are also part of contradiction and entailment pairs, and if negated sentences generally indicate negation, then the models would be expected to struggle to identify neutral samples that contain negation and that are also for humans somewhat difficult to identify as such.

Furthermore, as the hypotheses overlap almost entirely with the premises in our neutral samples, the biases observed by \citet{mccoy2019right} would be expected to contribute to this failure to identify the neutral samples as such.

One could wonder whether these results are of any significance for the models investigated. One could argue that it is no surprise that the models perform poorly on samples that differ relevantly from the ones that they have seen during training. Hence, it is to be expected -- and no reason for concern -- that the models fail to perform well at these tasks. 

In response to this, we would like to raise attention to the fact that the task is called natural language inference, and the concepts that the models are intended to learn are \emph{entailment} and \emph{contradiction}. If it turns out, as we suggest it has in our study and in similar ones referenced above, section 2, that the models' performance collapses even though the same logical relationships are in focus, while some superficial cues have disappeared, then it is wrong to say that the models are addressing the task of NLI, or that they are learning the logical concepts. Rather, the models are picking up spurious statistical cues that are correlated to the logical concepts in some dataset such as MNLI, but that are entirely unrelated to them in our dataset. 

In other words, we suggest that the current lack of generalization beyond the training dataset that we can observe in our study (but which is also more widely acknowledged, see the references in section 1 and 2) is indeed a reason for concern. It implies that the models do not actually learn NLI but rather the exploitation of spurious statistical cues in the dataset, leading to shallow heuristics.

\section{Conclusion}

So far, we have investigated the ability of state-of-the-art transformer-based PLMs fine-tuned on the common NLI datasets to recognize formal logical relationships in syllogistic patterns. Our results show that the models are very good at distinguishing entailment from contradiction, but very bad, much worse than chance, at distinguishing either from neutral. Our analysis has suggested that this is due to the PLMs' use of shallow heuristics, in particular with the attention to symmetries regarding negation and quantifiers between premises and hypotheses. We suggest that our study adds to the evidence that current transformer-based models do not actually learn NLI.

\bibliography{../../Bibliography/habil}

\newpage
\appendix

\section{Full Instructions Given to Crowdworkers}

\citet[1114]{Williams2018multinli} specifies the following tasks for the crowdworkers:

``This task will involve reading a line from a non-fiction article and writing three sentences that relate to it. The line will describe a situation or event. Using only this description and what you know about the world:
\begin{itemize}
	\item Write one sentence that is definitely correct about the situation or event in the line. 
	\item Write one sentence that might be correct about the situation or event in the line. 
	\item Write one sentence that is definitely incorrect about the situation or event in the line. ''
  
\end{itemize}

\section{Method used for evaluation of Models on MNLI}
\label{sec:app-evaluation}

To evaluate the models, we have used Huggingface's trainer API, see \href{https://huggingface.co}{Huggingface} \citep{Huggingface}. In particular, we followed the  instructions in the notebook \href{https://colab.research.google.com/github/huggingface/notebooks/blob/master/transformers\_doc/pytorch/training.ipynb}{here}. We evaluated the models using the API out-of-the-box, with the following exceptions:

\begin{enumerate}
  \item The textattack-models had as labels "LABEL\_0, LABEL\_1, LABEL\_2", which could not be read by the function that ensures that the labels are used equivalently by both model and dataset; hence, we reconfigured the models to use as labels ``contradiction, entailment, neutral''.
  \item facebook-bart-large-mnli by textattack posed two additional challenges. 
  \begin{enumerate}
    \item Due to out of memory issues, we had to split up processing of the validation set into three chunks, averaging the accuracy received afterwards.
  \item The logits containing the predictions issued by facebook-bart-large-mnli could not be processed by the evaluation function, which caused the need to select only the first slice of the tensor that the model was issuing, ensuring that the metric function got a 1-dimensional tensor to compute accuracy.
  \end{enumerate}
\end{enumerate}
\end{document}